# How do the naive Bayes classifier and the Support Vector Machine compare in their ability to forecast the Stock Exchange of Thailand?

Napas Udomsak, *Student, Bangkok Patana School*

*Abstract*—This essay investigates the question of how the naive Bayes classifier and the support vector machine compare in their ability to forecast the Stock Exchange of Thailand. The theory behind the SVM and the naive Bayes classifier is explored. The algorithms are trained using data from the month of January 2010, extracted from the MarketWatch.com website. Input features are selected based on previous studies of the SET100 Index. The Weka 3 software is used to create models from the labeled training data. Mean squared error and proportion of correctly classified instances, and a number of other error measurements are the used to compare the two algorithms. This essay shows that these two algorithms are currently not advanced enough to accurately model the stock exchange. Nevertheless, the naive Bayes is better than the support vector machine at predicting the Stock Exchange of Thailand.

## I. INTRODUCTION

Machine learning is a branch of artificial intelligence that is concerned with the construction of models from data (KOVAHI, Ron and Provost, Foster, 1998). In supervised machine learning, a subfield of machine learning, computers derive models from labeled training data. Recently, the field of machine learning has seen a rise in the popularity of probabilistic and statistical models. Notably, the Naive Bayes, Artificial Neural Networks (ANN), and Support Vector Machines (SVM).

The aim of supervised machine learning is that given a set of $N$ training examples, $\{(x_1, y_1), \ldots (x_n, y_n)\}$ where $x_i$ is the feature vector $i^{th}$ example, and $y_i$ is its label, or class, to derive a function $g: X \to Y$ mapping input feature $x$ to label $y$ such that $X$ is the input space $\{x_1, \ldots, x_n\}$ and $Y$ is the output space of possible classes $\{y_1, \ldots, y_n\}$ (MOHRI, Mehryar et al., 2012).

The prediction and forecasting of financial markets has been of interest to artificial intelligence researchers since the dawn of learning algorithms (OU, Phichhang and Wang, Hengshan, 2009). A stock market index is a statistical composite of the movement of the overall market. This index will reflect the performance of all companies in the stock market over a period of time.

As an economist and a computer scientist, investigating the stock market using computer algorithms is of great interest to me. Furthermore, if this experiment produces an accurate model a trading strategy could be created that would allow for profits to be derived from the buying and selling of stocks.

In this essay, I will be looking at the Stock Exchange of Thailand, specifically the SET100 index. Data from the month of January 2010 will be extracted and used to train a naive Bayes classifier and a support vector machine. I will then use a number of performance indicators to compare the algorithms. This will allow me to answer the question: How do the naive Bayes classifier and the support vector machine (SVM) compare in their ability to forecast the Stock Exchange of Thailand?

## II. ANALYSIS

### A. The Stock Exchange

The stock exchange is a place where buyers and sellers come together to trade shares. A stock market index measures the value of a section of the stock market. The SET100 is the primary index of the Stock Exchange of Thailand (SET). It tracks the prices of the top 100 companies on the SET ranked by market capitalization and liquidity. It can be calculated using the following formula (PHAISARN, Sutheebanjard and Wichian, Premchaiswadi, 2010):

$$\text{SET } 100 = \frac{\text{Current market value} \times 100}{\text{Base market value}}$$

### B. The Efficient Market Hypothesis (EMH)

The EMH was developed by Fama in 1970 (FAMA, Eugene F, 1970). It states that the price of a security reflects the complete market information. Should there be a change in financial outlook, the market is perfectly efficient, and will therefore instantly adjust the security price to reflect this new information. The EMH is controversial and often disputed. Its supporters claim that attempts at predicting stock price through technical or fundamental analysis is pointless as the market will immediately reflect any new information discovered. Hence, an abnormal profit cannot be obtained. Nevertheless, there are three different forms of the EMH. The weak EMH states that only historical information is embedded in the stock price. The semi-strong form claims that the price represents all historical information and all available public information, while the strong EMH suggests that the current stock price represents all available historical, public, and private insider information. Fama himself has considered the strong EMH to be invalid.

*C. Random walk theory*

A random walk is one in which the next step, or steps, cannot be predicted based on the information about past steps. In the context of a stock market, this theory proposes that the short-run direction of stock price cannot be predicted. Therefore investment services, earning predictions, and complicated chart patterns are all but a hoax. The stock's future prices take a completely random unpredictable path. In this theory technical and fundamental analysis are considered feeble attempts at trying to beat the market. Kendall and Hill first proposed this theory in 1953 and shocked many economists (KENDALL, Maurice George and Hill, Bradford A., 1953). However, after further debate and research, the formal random walk theory market was devised.

The random walk theory and the EMH are compatible. The market can move in unpredictable directions while being efficient. Both theories suggest that an abnormal profit cannot be achieved from stock price prediction

*D. Fundamental Analysis*

Fundamental analysis looks at the numerical indicators that describe the company underlying the stock e.g. P/E ratio (MCCLURE, Ben, 2014). An analysts looks at the fundamentals of a company in order to determine whether the stock is under, or over-valued and then buying, or selling the stock, respectively.

*E. Technical Analysis*

Technical analysts or 'chartists' are not concerned about the company's fundamentals. Technical analysis aims to derive patterns and trends from past price to predict future price (a form of time series analysis) (INTERACTIVE DATA CORP, 2014). Historical data are tested using specific rules for buying and selling in order to evaluate whether or not a profit can be made by following the same strategy in the future. Technical analysis is based on the assumption that the pattern derived from historical data will apply to future data.

*F. The naive Bayes classifier*

Bayesian classifiers are statistical classifiers that can predict the probability of a class membership, i.e. the probability that a given sample belongs to a specific class. Bayesian classifiers apply Bayes theorem in order to produce probabilities of class membership. The naive Bayes classifier, a special type of Bayesian classifier, utilizes a naive assumption of conditional independence between attributes. This assumption exists merely to simplify computation and is therefore "naive" (STUART, Russell J. and Norvig, Peter, 2009).

Suppose X is a single sample represented by a n-dimensional vector $\{x_1, x_2, \ldots, x_n\}$ that contains the n attributes of x. In Bayesian statistics, X is called the "evidence". Now, suppose H is a hypothesis that X belongs to some class C. In classification, we want to find the probability of $P(H|X)$ – the probability that X belongs to class C given the evidence $\{x_1, x_2, \ldots, x_n\}$. $P(H|X)$ is known as the posteriori probability of H given X.

The naive Bayes classifier works as follows:

Given a set of labelled training samples, T. Each item in T contains a label, $C_1$, $C_2$, …, $C_k$ for k possible classes. Each item in T also contains an n-dimensional vector $\{x_1, x_2, \ldots, x_n\}$ that represent the n attributes $\{A_1, A_2, \ldots, A_n\}$ which describe that item. Given a single sample X, the classifier will find the class $C_i$ that maximizes $P(C_i|X)$. In other words, X belongs to $C_i$ if and only if $P(C_i|X) > P(C_j|X)$ for $1 \leq j \leq k$ and $j \neq i$; Choosing a class based on this rule is also known as maximum-likelihood estimation (WAZIRI, Victor Onomza, 2013).

By Bayes theorem (INTERNATIONAL BACCALAUREATE, 2012), $P(C_i|X)$ can also be calculated using the following formula:

$$P(C_i|X) = \frac{P(X|C_i)\,P(C_i)}{P(X)}$$

Since P(X) is equal for all classes, $P(C_i|X)$ is maximized when $P(X|C_i)P(C_i)$ is maximized. If $P(C_i)$ is not known, it is common to assume a uniform distribution for $P(C_i)$ such that $P(C_1) = P(C_2) = \ldots = P(C_k)$. Hence, $P(C_i|X) \propto P(X|C_i)$ and our goal becomes to maximize $P(X|C_i)$. Priori probability, $P(C_i)$, can also be estimated by the following approximation:

$$P(C_i) = \frac{\text{freq}(C_i \text{ in } T)}{\text{size}(T)}$$

where T is the training set used to create the classifier. Because it is computationally expensive to calculate $P(X|C_i)$, the naïve assumption of conditional independence is utilized to allow us to make the following approximation:

$$P(X|C_i) = P(x_1, x_2, \ldots, x_n|C_i) \cong \prod_{k=1}^{n} P(x_k|C_i)$$

$P(x_1|C_i)$, $P(x_2|C_i)$, …, $P(x_n|C_i)$ can now be easily calculated from the labelled training set.

Recall that xk represents the measured value of an attribute Ak. If Ak is categorical, then the P(xk|Ci) of can be calculated by counting the occurrence of xk in samples labelled Ci divided by the number of Ci examples in T, i.e.:

$$P(x_i|C_i) = \frac{\text{freq}(C_i \text{ containing } x_i \text{ in } T)}{\text{freq}(C_i \text{ in } T)}$$

However, if Ak is continuous then we can assume that it is approximated by a Gaussian distribution (INTERNATIONAL BACCALAUREATE, 2012) with a mean, μ, and a standard deviation, σ, defined by:

$$g(x, \mu, \sigma) = \frac{1}{\sqrt{2\pi}\sigma} e^{-\frac{(x-\mu)^2}{2\sigma^2}}$$

where μ and σ can be estimated by:

$$\mu = \frac{\sum_{k=1}^{n} x_k \text{ in } C_i}{n}$$

$$\sigma = \frac{\sum_{k=1}^{n} x_k^2 \text{ in } C_i}{n} - \left(\frac{\sum_{k=1}^{n} x_k \text{ in } C_i}{n}\right)^2$$

with this we can calculate now calculate P(xk|Ci):

$$P(x_k|C_i) = g(x_k, \mu_{C_i}, \sigma_{C_i})$$

*G. Laplace Smoothing*

In the event that Ak is categorical and xk does not fit within the bounds of any category of Ak such that P(xk|Ci) = 0 then P(X|Ci) = 0 because recall that P(X|Ci) is calculated by $P(X|C_i) \cong \prod_{k=1}^{n} P(x_k|C_k)$. Hence, to prevent this error, the Laplacian correction can be utilized to deal with zero probability values (PRABHAKAR, Raghavan et al., 2008).

With the Laplacian estimator, suppose that there is a sample X that contains feature xk that describes the attribute Ak of X. If xk does not fit within the bounds of any category then instead of assigning the probability P(xk|Ci) = 0, we use the below formula to estimate the probability of P(xk|Ci).

$$P(x_k|C_i) = \frac{1}{\text{freq}(x_k \text{ in } T)}$$

*H. Support Vector Machines*

Support vector machines (SVM) are another tool used in supervised machine learning to classify a sample X to a class labels C. SVMs work by deriving a linear decision boundary that can represent a non-linear class boundary through a non-linear mapping of input vectors xk into a higher-dimensional feature space. The linear model is constructed in the higher-dimensional feature space to represent a non-linear decision boundary (WESTON, Jason, 2004).

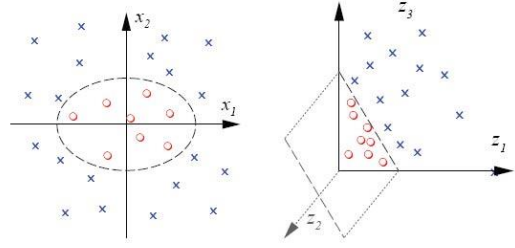

Figure 1. (THORNTON, Chris, 2010)
Left: Non-linear decision boundary in 2 dimensions
Right: Linear decision boundary in 3 dimensions

In the new space a maximum margin hyper-plane is derived from training data. This maximum margin hyper-plane provides maximum separation between two classes. This hyper-plane is derived from the examples closest to it, all other examples are considered irrelevant in defining the decision boundary.

For the linearly separable case where there two classes and the data is represented by three attributes x1, x2, x3, there is no need to map to a higher-dimensional space and thus the maximum margin hyper-plane will have an equation of the following form:

$$y = w_0 + w_1 x_1 + w_2 x_2 + w_3 x_3$$

where y is the outcome, xi, are the attributes. The four weights, wi, are learned from the training data.

The maximum margin hyper-plane can also be represented in terms of the support vectors.

$$y = b + \sum \alpha_i y_i x(i) \cdot x$$

where yi is the class outcome of the specific training example x(i) and the · is dot product. The vector x represents a test example and x(i) are the support vectors that are used to determine the decision boundary. In this equation, αi and b are parameters that are optimized in the process of finding the maximum margin hyper plane. At implementation this turns in to a linearly constrained quadratic programming problem whereby the support vectors x(i) are found, and parameters, b and αi , are determined (KIM, Kyoung-jae, 2003).

For the nonlinearly separable case where the input must be mapped to a higher-dimensional feature space, the decision boundary can be represented as follows:

$$y = b + \sum \alpha_i y_i K(x(i), x)$$

where K(x(i),x) is defined as the kernel function.

There are a number kernel functions that we can choose from. These functions define how the SVM performs the mapping of input features to a higher dimensional space.

Common kernel functions include the polynomial kernel $K(x,y) = (xy+1)^d$ where d is the degree of the polynomial and the Gaussian radial basis function $K(x,y) = \exp(-1/\delta^2 (x-y)^2)$ where $\delta^2$ is the bandwidth of the radial basis function. $\delta^2$ is usually selected via a grid search (HUANG, Wei et al., 2005).

*I. Advantages and Disadvantages of the naive Bayes classifier and SVM*

The naive Bayes classifier is easy to implement because we can easily produce probability estimates using the formulas detailed in the section 2.4. Studies have shown that the naive Bayes can produce good parameter estimates with small data sets (UDDIN, Ashraf et al., 2012).

On the other hand, the main disadvantage of this classifier is its naive assumption of conditional independence. In practice, dependencies exist between variables, especially in a complicated system like the stock market. These dependencies are ignored by the naive Bayes classifier which would cause it to produce less accurate predictions (ESMAEL, Bilal, 2013).

The maximum margin decision boundary utilized by the SVM is built upon a unique principle of structural risk minimization whereby the decision boundary is derived through minimizing the upper bound of the generalization error, allowing SVMs to be very resilient against the over-fitting problem. Because the optimization of parameters in an SVM can be achieved through solving a linearly constrained quadratic programming problem, the solution will always be a unique global optimum (HUANG, Wei et al., 2005).

SVMs require relatively complicated high-dimensional models that take a longer time to train when compared to the naive Bayes. Their method of mapping input features into higher-dimensional spaces is controlled by a kernel function has to be manually selected. The selection of an inappropriate kernel function can heavily detriment the performance of a SVM (CHRISTOPHER, Burges JC., 1998). Parameter tuning of the SVM must also be performed manually using a grid search. Once again, in appropriate selection of the sigma parameter of the radial basis kernel can lead to issues like over fitting. Research has shown that when dealing with extremely large data sets SVMs result in high algorithmic complexity and extensive memory requirements (HORVATH, Gabor et al., 2003).

One algorithm is not categorically better than the other. It is difficult to predict the real-world performance of learning algorithms based on their theoretical advantages and disadvantages, as each data set has its own unique patterns.

### III. EXPERIMENT

*A. Hypothesis*

Based on the theory discussed above, I believe that a support vector machine will produce better predictions of the stock market direction than a naive Bayes classifier. The SVMs ability to model non-linear decision boundaries using a linear model will be advantageous as the stock market is most definitely a complicated nonlinear process. The naive assumption of conditional independence, that is central to the naive Bayes classifier, will be penalizing to its performance. Because of the inter-connected nature of the stock market, it is unlikely that a change in one factor will be completely independent to a change in another factor in real life.

*B. Hardware and software specifications*
CPU: Intel Core i5-3337U 1.80Ghz
RAM: 4.00 GB
OS: Windows 8 x64
Required software: Java SE Runtime Environment 7 and Weka 3

*C. Method*
Input features:
1. Nikkei 255 Index (NK)
2. Hang Seng Index (HS)
3. SET 100 Index (SET)
4. USDTHB Exchange Rate (USDTHB)
5. S&P 500 Index (SP)
6. COMEX Gold Futures (GOLD)

These variables have been chosen from previous studies on factors that affect stock market direction (SUTHEEBANJARD, Phaisarn and Premchaiswadi, Wichian, 2009). External factors such as the index have been chosen to represent the market sentiment, whereas internal factors like the SET index itself should tell the classifier about the internal situation of the market.

First, the data of the input variables will be extracted for the period 1st January 2010 – 1st February 2010 from MarketWatch.com (see Appendix A on pg.18 for program code used to extract data).

Secondly, the data will be processed and the percentage change between each day will be found.

Thirdly, the stock market direction will be classified either UP or DOWN (this becomes the class label). Hence, a single sample, X, in the training set T will look as follows (see Appendix B on pg.22 for full training set):

| NK | HS | SET | USDTHB | SP500 | GOLD | SET_DIRECTION |
|---|---|---|---|---|---|---|
| 0.69941 | 0.206943 | -0.37647 | 0.153233 | -1.105 | -0.46795 | UP |

Fourthly, create a naive Bayes Classifier and a SVM from the data using the WEKA software (see Appendix C pg.23 for step-by-step guide)

Finally, calculate indicators of classifier performance over 10-fold cross validation

IV. RESULTS

|  | SVM | Naive Bayes |
|---|---|---|
| Correctly classified instances | 56 % | 66 % |
| Mean absolute error | 0.43 | 0.38 |
| Root mean squared error | 0.65 | 0.54 |
| Relative absolute error | 86 % | 77 % |
| Root relative squared error | 130 % | 108 % |

*data is based on naive Bayes trained on continuous values (see Appendix D for parameters of Gaussian fitting) and SVM parameter $\delta 2 = 0$

Confusion matrix for SVM:
$$\begin{pmatrix} a & b & \leftarrow \text{classified as} \\ 11 & 5 & a = \text{up} \\ 8 & 6 & b = \text{down} \end{pmatrix}$$

Confusion matrix for naive Bayes:
$$\begin{pmatrix} a & b & \leftarrow \text{classified as} \\ 13 & 3 & a = \text{up} \\ 7 & 7 & b = \text{down} \end{pmatrix}$$

*A. Evaluation*

During the experiment a limited amount of data was chosen to train these models, a larger training data set will produce different models that might be better. The method of testing chosen was 10-fold cross validation, while this helps analyse the extent of over-fitting, the model is still ultimately still being tested on its own training data. A completely independent data set could have been used to produce better indicators of performance. The data is assumed to be continuous in the training of models, some studies have shown that discretization could be beneficial to the performance of these models, especially the naive Bayes classifier. In future experiments, I could discretize the data before using them to create models.

*B. Conclusion*

The results of 10-fold stratified cross validation show that the naive Bayes displays better performance in terms of the proportion of correctly classified instances, and lower error on every measurement of error. This does not support my hypothesis. I believe this could be due to the SVM's over complexity and the Gaussian smoothing that the naive Bayes classifier was able to benefit from.

Ultimately, both models were able to predict the stock market to some degree. While, these results cannot be used to support or dispute the EMH as the data is all in the past, it does suggest that perhaps the stock market does not follow a random walk. It is probable that behind all the noise and chaos of market there is a complex non-linear process, however today's algorithms are clearly not yet advanced enough to accurately model such a process.

Future research should consider resolving the conditional independence assumption by modelling the relationships between variables. I believe research of more advanced kernel functions will be the key to improving the performance of the support vector machine.

In conclusion, the naive Bayes was able to forecast the stock market better than the SVM based on data from the month of January 2010.

International Journal of Advanced Computer Technology. II(1).

WESTON, Jason. 2004. Support Vector Machine (and Statistical Learning Theory) Tutorial. [online]. [Accessed 10 November 2014]. Available from World Wide Web: <http://www.cs.columbia.edu/~kathy/cs4701/documents/jason_svm_tutorial.pdf>

The below code extracts data from www.marketwatch.com

```java
import java.io.FileWriter;
import java.text.SimpleDateFormat;
import java.util.Arrays;
import java.util.Calendar;
import java.util.Date;
import org.jsoup.Jsoup;
import org.jsoup.nodes.Document;
import org.jsoup.select.Elements;

import au.com.bytecode.opencsv.CSVWriter;

public class DataExtraction {

    private static final String NK_LINK = "http://www.marketwatch.com/investing/index/nik/historical?CountryCode=jp";
    private static final String HS_LINK = "http://www.marketwatch.com/investing/Index/HSI/historical?CountryCode=HK";
    private static final String SET_LINK = "http://www.marketwatch.com/investing/index/set/historical?CountryCode=th";
    private static final String USDTHB_LINK = "N/A";
    private static final String SP500_LINK = "http://www.marketwatch.com/investing/index/SPX/historical";
    private static final String GOLD_LINK = "http://www.marketwatch.com/investing/future/gold/historical";
    private static final String LINKS[] = new String[] { NK_LINK, HS_LINK,
            SET_LINK, USDTHB_LINK, SP500_LINK, GOLD_LINK };
    private static final SimpleDateFormat sdf = new SimpleDateFormat(
            "dd MMM yyyy");

    public static void main(String[] args) throws Exception {

        Calendar cal = Calendar.getInstance();
        cal.set(2010, 0, 1);

        CSVWriter raw = new CSVWriter(new FileWriter("Rawdata.csv", true));

        raw.writeNext("TIME,NK_CLOSE,HS_CLOSE,SET_CLOSE,USDTHB_CLOSE,SP500_CLOSE,GOLD_CLOSE,MLR,SET_OPEN"
                .split(","));

        CSVWriter processed = new CSVWriter(new FileWriter("Processed.csv",
                true));
        processed
                .writeNext("TIME,NK_CHANGE,HS_CHANGE,SET_CHANGE,USDTHB_CHANGE,SP500_CHANGE,GOLD_CHANGE,MLR,TODAY_SET_DIRECTION"
                        .split(","));

        CSVWriter allData = new CSVWriter(new FileWriter("allData.csv", true));
        processed
                .writeNext("TIME,NK_CHANGE,HS_CHANGE,SET_CHANGE,USDTHB_CHANGE,SP500_CHANGE,GOLD_CHANGE,MLR,TODAY_SET_DIRECTION"
                        .split(","));

        String[] today, ytd = new String[] { "0", "0", "0", "0", "0", "0", "0",
                "0" }, twoDaysAgo = new String[] { "0", "0", "0", "0", "0",
                "0", "0", "0" }, calcData, SET_Today;
```

```java
for (int i = 1; !sdf.format(cal.getTime()).equalsIgnoreCase(
        "31 Jan 2010"); i++) {

    long start_time = System.currentTimeMillis();

    today = new String[9];
    calcData = new String[9];

    today[0] = sdf.format(cal.getTime());

    SET_Today = getMarketWatchData(SET_LINK, cal.getTime());

    for (int j = 0; j < 6; j++) {
        if (j == 3) {
            today[j + 1] = getUsdThbData(cal.getTime());
            continue;
        }
        String[] data = getMarketWatchData(LINKS[j], cal.getTime());
        if (data != null) {
            today[j + 1] = data[0];
        } else {
            break;
        }
    }

    // check for null and replace commas
    boolean noNull = true;
    for (int j = 0; j < today.length; j++) {
        if (today[j] == null) {
            noNull = false;
            break;
        } else {
            today[j] = today[j].replace(",", "");
        }
    }

    if (noNull) {
        calcData[0] = sdf.format(cal.getTime());
        for (int j = 1; j <= 6; j++) {
            calcData[j] = Double.toString((100 * (Double
                    .parseDouble(ytd[j]) - Double
                    .parseDouble(twoDaysAgo[j])))
                    / Double.parseDouble(twoDaysAgo[j]));
        }
        calcData[7] = ytd[7];

        SET_Today[0] = SET_Today[0].replace(",", "");
        SET_Today[1] = SET_Today[1].replace(",", "");
        calcData[8] = Double.parseDouble(SET_Today[1]) < Double
                .parseDouble(SET_Today[0]) ? "UP" : "DOWN";

        raw.writeNext(today);
        raw.flush();

        processed.writeNext(calcData);
        processed.flush();
```

```java
						System.out.println(cal.getTime().toString() + ": "
								+ Arrays.toString(today));

						twoDaysAgo = ytd;
						ytd = today;

				} else {
						System.out.println(cal.getTime().toString() + ": "
								+ Arrays.toString(today));
				}

				cal.add(Calendar.DAY_OF_YEAR, 1);

				System.out.println("Time taken: "
						+ (System.currentTimeMillis() - start_time) / 1000);

				allData.writeNext(today);
				allData.flush();
		}

		raw.close();
		processed.close();
		allData.close();
}

// USDTHB exchange rate
static String getUsdThbData(Date date) {
		SimpleDateFormat ft = new SimpleDateFormat("M-d-yyyy");

		try {

				Document doc = Jsoup.connect(
								"http://www.exchange-rates.org/Rate/THB/USD/"
										+ ft.format(date)).get();
				Elements data = doc.select("span#ctl00_M_grid_ctl02_lblResult");
				return data.get(0).text()
								.substring(0, data.get(0).text().length() - 4);

		} catch (Exception e) {
		}
		return null;
}

// MarketWatch [open, close] price
static String[] getMarketWatchData(String marketWatchLink, Date date) {
		SimpleDateFormat ft = new SimpleDateFormat("MM/dd/yyyy");
		try {

				Document doc = Jsoup.connect(marketWatchLink)
								.data(new String[] { "date", ft.format(date) }).post();

				Elements data = doc.select("img[src^=/investing/investing/]");

				String s;
				String out[] = new String[2];
				for (int i = 0; i < 2; i++) {
						s = data.get(i).toString();
```

```
                        out[i] = new StringBuffer(s.substring(s.indexOf("data=") + 5,
                                        s.indexOf("\"", s.indexOf("data=")))).reverse()
                                        .toString();
                }
                return out;
        } catch (Exception e) {
        }
        return null;
    }

}
```

APPENDIX B

| NK | HS | SET | USDTHB | SP500 | GOLD | SET_DIRECTION |
|---|---|---|---|---|---|---|
| 0.6994 | 0.2069 | -0.3765 | 0.1532 | -1.1050 | -0.4680 | UP |
| 2.8179 | 2.9316 | 1.7767 | 0.3672 | 4.5664 | -0.4345 | DOWN |
| -0.8336 | -0.0057 | -0.0946 | 0.2134 | -0.3123 | -0.1973 | UP |
| -0.8581 | -0.9368 | 0.1420 | 0.0000 | -0.3242 | 0.9642 | UP |
| 0.6710 | 0.4642 | 0.4325 | 0.1217 | 0.2656 | -0.3974 | DOWN |
| 0.7002 | 0.5851 | -1.2273 | 0.3342 | 0.7597 | 1.3817 | UP |
| 1.3981 | -0.3864 | 0.4317 | 0.0000 | -0.0048 | -1.0339 | DOWN |
| 0.7176 | 0.5048 | 0.7648 | 1.0902 | 0.0197 | 1.3712 | DOWN |
| -2.5613 | -0.1049 | -0.4723 | 0.3595 | 0.0197 | -0.0410 | DOWN |
| 0.0868 | -0.0738 | 0.3397 | 0.1493 | 0.5643 | 0.4803 | UP |
| 2.8621 | 1.1226 | 0.9494 | 0.0596 | 0.3403 | -0.2215 | DOWN |
| -1.8635 | 0.2424 | -0.0244 | 0.1787 | 0.4428 | 0.3564 | DOWN |
| -2.0816 | -0.1010 | 0.3563 | -0.0595 | 0.1507 | -0.3843 | UP |
| 1.2766 | -0.1532 | 0.6872 | -0.1488 | 0.0007 | -0.9819 | UP |
| 2.8790 | -0.0783 | 0.8502 | -0.4470 | 0.5445 | -0.7614 | UP |
| -0.9366 | 0.3878 | 0.7281 | 0.0898 | -0.1850 | -0.2320 | UP |
| 0.3918 | -0.0706 | 0.4565 | 0.0598 | 0.5106 | 0.4829 | UP |
| 2.2751 | 0.7055 | 0.8149 | 0.3586 | -0.3900 | 1.1332 | DOWN |
| 0.2223 | -0.3884 | -1.1148 | -0.2382 | -0.2563 | -1.1792 | UP |
| 0.4729 | -0.0324 | 1.6972 | 0.1791 | 1.0053 | 0.4927 | DOWN |
| 0.6166 | -0.1553 | 0.4769 | 0.0894 | -1.1539 | 0.3367 | UP |
| -1.8954 | -2.2651 | -0.0432 | 0.1191 | 1.0416 | -0.1649 | DOWN |
| 3.7733 | 0.4683 | -0.3566 | -0.1784 | 0.0549 | 0.2949 | UP |
| -0.9306 | -0.3437 | -0.0360 | 0.0000 | -0.1805 | -0.4528 | DOWN |
| -1.7955 | 0.1646 | -0.1674 | -0.0298 | 0.5658 | -0.2540 | UP |
| 1.3819 | 0.8533 | 1.9662 | -0.1192 | 0.2273 | -1.9838 | DOWN |
| -1.1802 | 0.1337 | -0.3419 | -0.1790 | -0.1045 | -1.5829 | UP |
| 2.6352 | -0.5850 | 1.6510 | 0.1195 | -0.5159 | -1.9583 | DOWN |
| -1.3877 | -1.7194 | -1.1574 | 0.0896 | -0.6303 | -0.0188 | UP |
| 0.6792 | -0.5423 | 0.7451 | 0.0298 | 0.8773 | -0.4071 | DOWN |

APPENDIX C

Guide for using Weka (Waikato Environment for Knowledge Analysis) 3 to create a naive Bayes classifier and SVM.

1. Start the Weka explorer

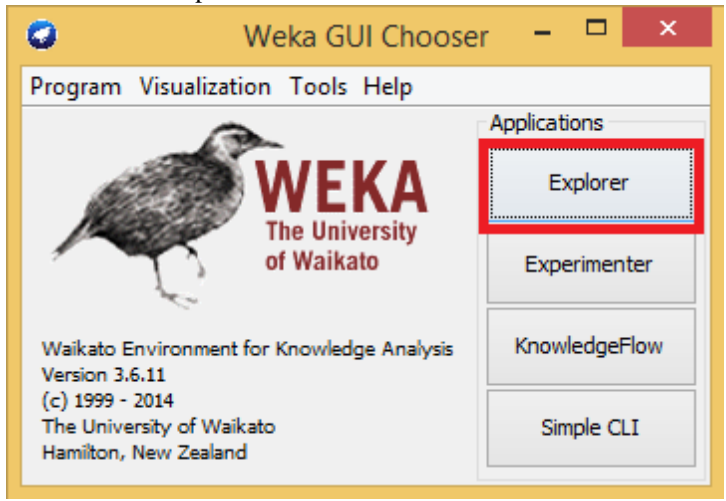

2. Select training data

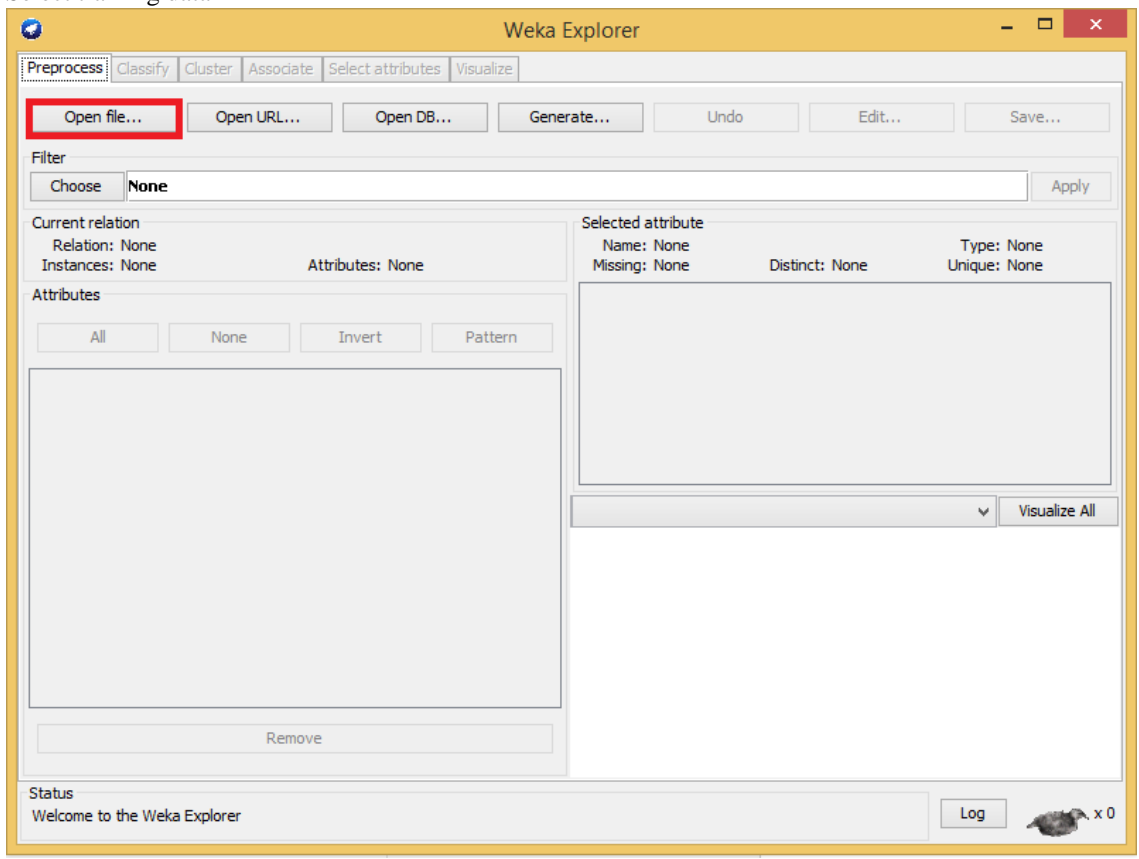

3. Switch to the "classify" tab

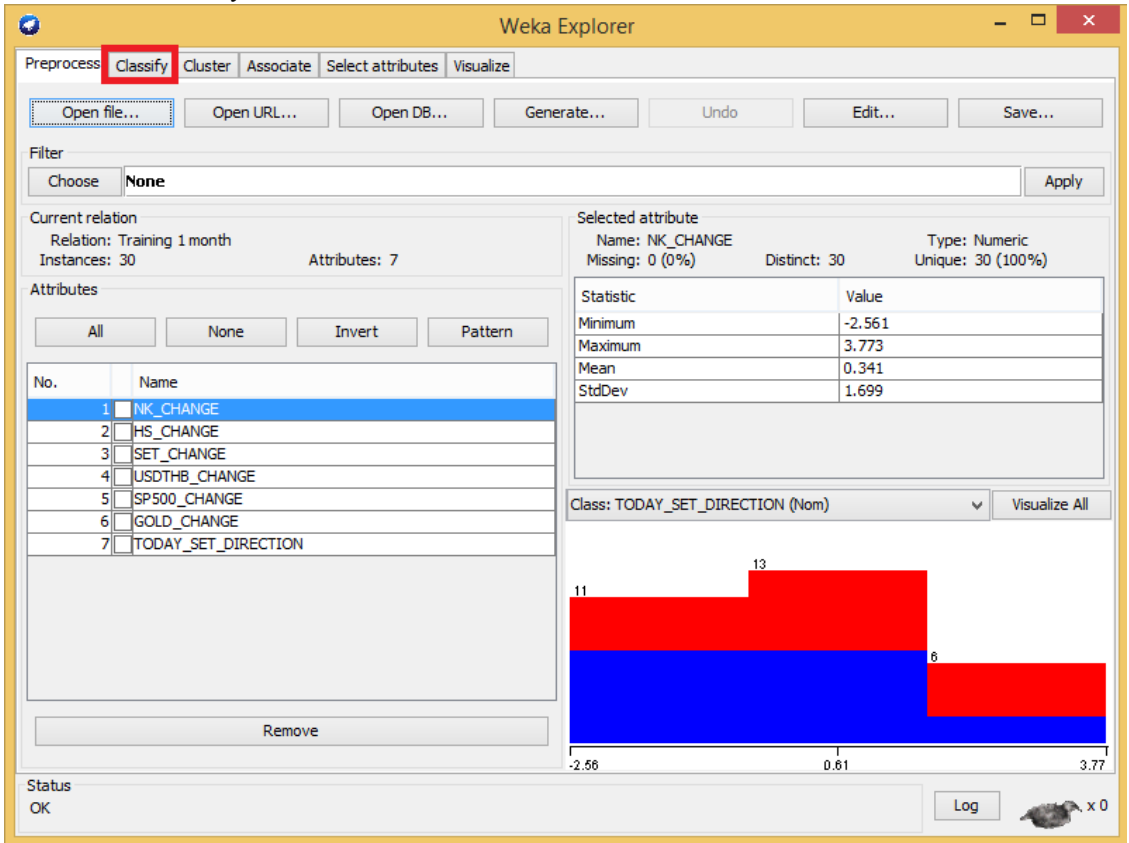

4. Choose classifier, cross-validation, and label to be predicted or learned, and then click start

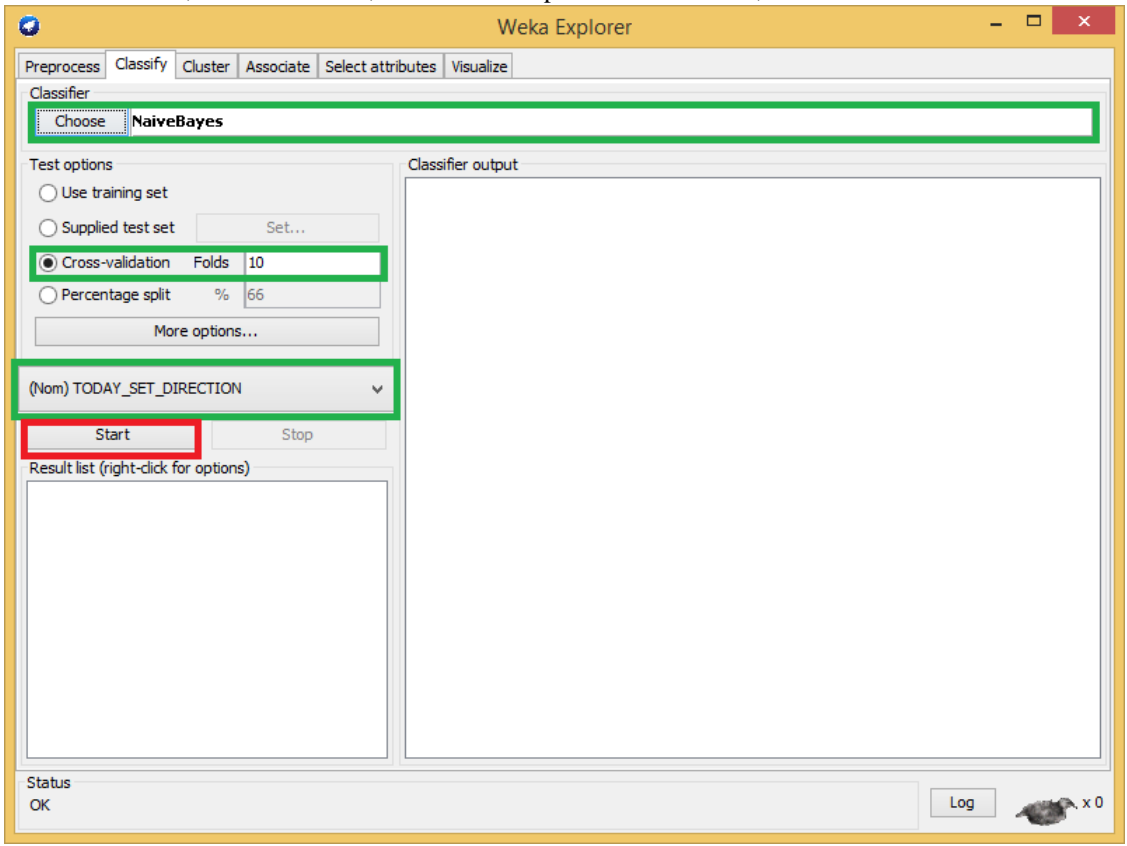

5. Extract results from classifier output

```
=== Stratified cross-validation ===
=== Summary ===

Correctly Classified Instances         20               66.6667 %
Incorrectly Classified Instances       10               33.3333 %
Kappa statistic                         0.3182
Mean absolute error                     0.3893
Root mean squared error                 0.5449
Relative absolute error                77.3338 %
Root relative squared error           107.9757 %
Total Number of Instances              30

=== Detailed Accuracy By Class ===

                 TP Rate  FP Rate  Precision  Recall  F-Measure  ROC Area
                 0.813    0.5      0.65       0.813   0.722      0.612
                 0.5      0.188    0.7        0.5     0.583      0.612
Weighted Avg.    0.667    0.354    0.673      0.667   0.657      0.612

=== Confusion Matrix ===

  a  b   <-- classified as
 13  3 |  a = UP
```

APPENDIX D

Gaussian parameters used in the naive Bayes classifier

| Attribute | UP | DOWN |
| --- | ---: | ---: |
| NK μ | 0.0956 | 0.5773 |
| NK σ | 1.5406 | 1.7603 |
| HS μ | -0.1008 | 0.1792 |
| HS σ | 0.5522 | 1.1067 |
| SET μ | -0.0551 | 0.7708 |
| SET σ | 0.6573 | 0.7328 |
| USDTHB μ | 0.0000 | 0.1993 |
| USDTHB σ | 0.1972 | 0.2779 |
| SP500 μ | -0.0616 | 0.5213 |
| SP500 σ | 0.5906 | 1.2035 |
| GOLD μ | -0.1306 | -0.2653 |
| GOLD σ | 0.7586 | 0.9391 |